\documentclass[10pt,twocolumn,letterpaper]{article}

\usepackage{iccv}
\usepackage{times}
\usepackage{epsfig}
\usepackage{graphicx}
\usepackage{amsmath}
\usepackage{amssymb}

\usepackage{booktabs}
\usepackage{bm}
\usepackage{amsfonts}
\usepackage{multirow}
\usepackage{makecell}
\usepackage{microtype}

\newcommand{\darrow}{$\downarrow$}
\newcommand{\uarrow}{$\uparrow$}
\newcommand{\tbf}[1]{\textbf{#1}}
\newcommand{\tul}[1]{\underline{#1}}

\usepackage{scalerel}
\newlength\bshft
\def\fakebold#1{\ThisStyle{\ooalign{$\SavedStyle#1$\cr%
  \kern-\bshft$\SavedStyle#1$\cr%
  \kern\bshft$\SavedStyle#1$}}}

\newcommand{\ET}{$\mathbb{ET}$}
\newcommand{\ETbold}{$\protect\fakebold{\mathbb{ET}}$}
\newcommand{\ETtitle}{\texorpdfstring{\ETbold}{ET}}

\usepackage[nocompress]{cite}

\usepackage[pagebackref=true,breaklinks=true,letterpaper=true,colorlinks,bookmarks=false]{hyperref}

\usepackage[capitalize]{cleveref}
\crefname{section}{Sec.}{Secs.}
\Crefname{section}{Section}{Sections}
\Crefname{table}{Table}{Tables}
\crefname{table}{Tab.}{Tabs.}

\iccvfinalcopy 

\def\iccvPaperID{3875} 

\ificcvfinal\fi

\begin{document}

\title{EigenTrajectory: Low-Rank Descriptors for Multi-Modal Trajectory Forecasting}

\author{Inhwan Bae\textsuperscript{\rm 1}\thanks{Work done during an internship at Carnegie Mellon University.} \qquad\qquad %
Jean Oh\textsuperscript{\rm 2} \qquad\qquad %
Hae-Gon Jeon\textsuperscript{\rm 1}\thanks{Corresponding author}\\
\textsuperscript{\rm 1}GIST AI Graduate School \qquad \textsuperscript{\rm 2}Carnegie Mellon University\\
{\tt\small inhwanbae@gm.gist.ac.kr, jeanoh@cmu.edu, haegonj@gist.ac.kr}
}

\maketitle
\ificcvfinal\thispagestyle{empty}\fi

\begin{abstract}
Capturing high-dimensional social interactions and feasible futures is essential for predicting trajectories. To address this complex nature, several attempts have been devoted to reducing the dimensionality of the output variables via parametric curve fitting such as the Bézier curve and B-spline function. However, these functions, which originate in computer graphics fields, are not suitable to account for socially acceptable human dynamics. In this paper, we present EigenTrajectory (\ET), a trajectory prediction approach that uses a novel trajectory descriptor to form a compact space, known here as \ET~space, in place of Euclidean space, for representing pedestrian movements. We first reduce the complexity of the trajectory descriptor via a low-rank approximation. We transform the pedestrians' history paths into our \ET~space represented by spatio-temporal principle components, and feed them into off-the-shelf trajectory forecasting models. The inputs and outputs of the models as well as social interactions are all gathered and aggregated in the corresponding \ET~space. Lastly, we propose a trajectory anchor-based refinement method to cover all possible futures in the proposed \ET~space. Extensive experiments demonstrate that our EigenTrajectory predictor can significantly improve both the prediction accuracy and reliability of existing trajectory forecasting models on public benchmarks, indicating that the proposed descriptor is suited to represent pedestrian behaviors.
Code is publicly available at \url{https://github.com/inhwanbae/EigenTrajectory}.
\end{abstract}

\section{Introduction}
Trajectory prediction involves forecasting the future footsteps of agents based on their past movements. This task is considered one of the core technologies for autonomous navigation, social robot platforms and surveillance systems.

Many existing approaches~\cite{alahi2016social,gupta2018social,zhou2012understanding,zhou2015learning,li2019conditional,fernando2018gdgan,shi2020multimodal,kosaraju2019social,sun2020reciprocal,salzmann2020trajectron++,mohamed2020social,liang2020garden,Shi2021sgcn,yu2020spatio,li2020Evolvegraph,mangalam2020pecnet,bae2022npsn,bae2021dmrgcn,liu2021causal,li2021stcnet} design their prediction models in the Euclidean space, \ie, to directly infer a sequence of 2D coordinates of future frames. These approaches force the models to learn both informative behavioral features and their motion dynamics from raw trajectory data. Such direct predictions can intuitively describe agents' behaviors in the temporal series of spatial coordinates; however, in a higher-dimensional space, it is hard for the models to determine explanatory features.

Recent works have described the pedestrian's movements using trajectory descriptors instead of dealing with all raw coordinate information. Inspired by human beings traveling pathways with a higher level of connotation (\eg, a person who gradually decelerates to turn right, or makes a sharp turn while going straight)~\cite{rudenko2020survey}, parametric curve functions are introduced. In particular, Hug \etal~\cite{hug2020bezier,hug2022bezier} introduce the Bézier curve, and Jazayeri and Jahangiri~\cite{jazayeri2022bspline} propose a B-spline curve-based representation for effectively modeling continuous-time trajectories. These methods successfully reduce the dimensionality of trajectories by abstractly representing lengthy sequential coordinates in a spatial domain using a smaller set of key points. 
It is, however, unclear how well these parametric functions can capture human motions and behaviors as they have been designed for 
computer graphics~\cite{fey2018splinecnn,sharma2020parsenet,peng2017parametric} and part modeling~\cite{willis2021engineering,wu2021deepcad,wang2020pienet}.

\begin{figure}[t]
\centering
\includegraphics[width=\linewidth,trim={0 0 0 0},clip]{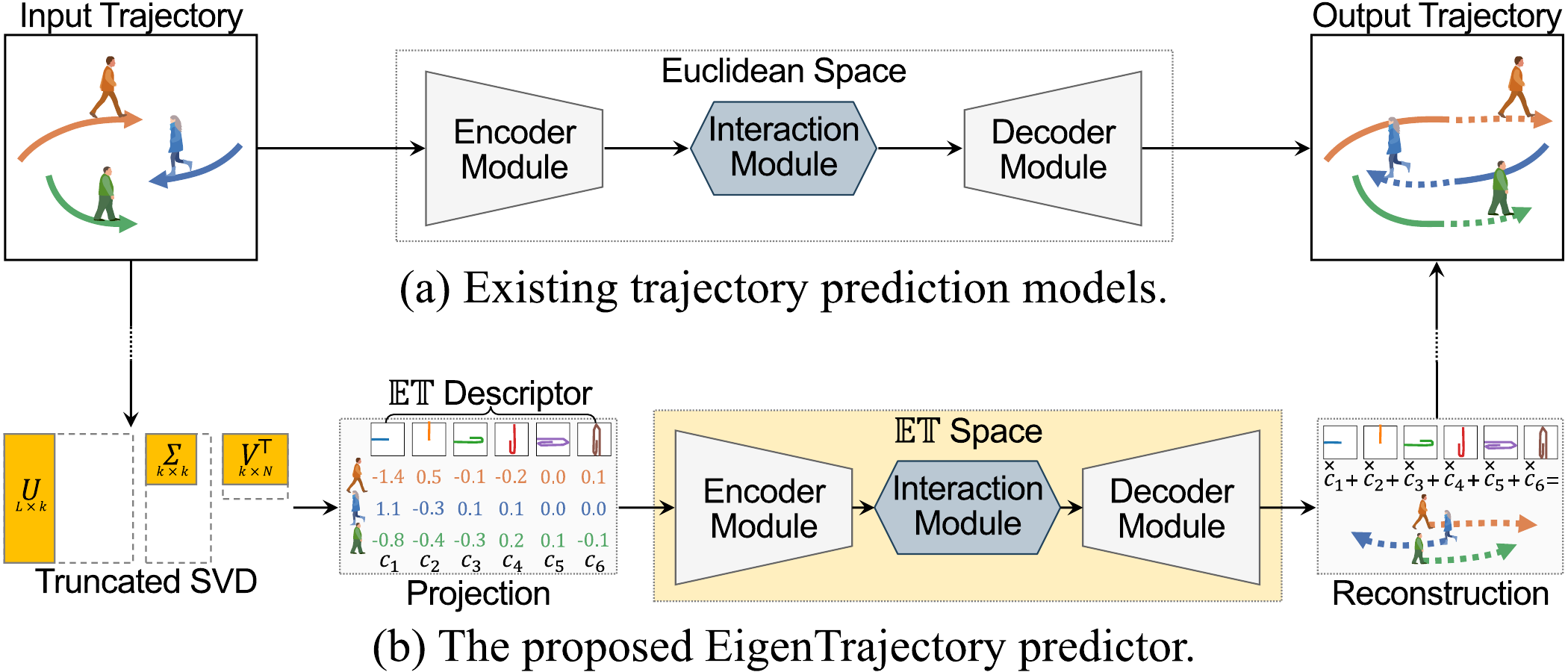}
\vspace{-5mm}
\caption{A common pipeline of trajectory prediction models and the proposed EigenTrajectory. For each observation, (a) existing models predict future trajectories using the raw data in Euclidean space; (b) our approach transforms the raw data into our \ET~space, then captures the social interaction and predicts the coefficient of our trajectory descriptor.}
\label{fig:teaser_model}
\vspace{-1mm}
\end{figure}

In this paper, departing from existing parametric curve functions, we present an intuitive trajectory descriptor that is learnable from real-world human trajectory data as illustrated in~\cref{fig:teaser_model}. First, we decompose a stacked trajectory sequence using Singular Value Decomposition (SVD). To represent the data concisely, we reduce the dimensionality by performing the best rank-$k$ approximation. Through this process, all trajectories can be approximated as a linear combination of $k$ eigenvectors, which we call EigenTrajectory (\ET) descriptor. Next, we aggregate the social interaction features and then predict the coefficients of the eigenvectors for the future path in the same space. Here, after clustering the coefficients, a set of trajectory anchors, which can be interpreted as the coefficient candidates, is used to ensure a diversity of prediction paths. Finally, trajectory coordinates can be reconstructed through the matrix multiplication of the eigenvector and the coefficients. Experimental results demonstrate that the proposed EigenTrajectory (\ET) descriptor can successfully represent the pedestrian motion dynamics and significantly improve the prediction accuracy of existing prediction models on various public benchmarks.

\section{Related Works}
\subsection{Pedestrian Trajectory Prediction}
Pioneering works~\cite{helbing1995social,pellegrini2009you,mehram2009socialforcemodel,yamaguchi2011you} model pedestrian movements following the social force theory or energy minimization to describe surrounding interactions and future movements. Since then, there have been significant advances in both social interaction and motion modeling with the introduction of Convolutional Neural Networks (CNNs) and Recurrent neural networks (RNNs) for trajectory prediction. One pioneering work is Social-LSTM~\cite{alahi2016social}, which introduces a long short-term memory (LSTM) to recurrently predict future coordinates. Agent's LSTM hidden states are shared with each other by a social pooling mechanism which leverages a neighbor's hidden state information inside a spatial grid. Attention mechanisms~\cite{vemula2018social,ivanovic2019trajectron,fernando2018soft,salzmann2020trajectron++} are also used to directly share social relations based on neighbors' influence. Better prediction results are achievable when using additional surveillance view images~\cite{varshneya2017,xue2018sslstm,manh2018scene,sadeghian2019sophie,liang2019peeking,kosaraju2019social,sun2020reciprocal,dendorfer2020goalgan,dendorfer2021mggan,zhao2019matf,tao2020dynamic,sun2020rsbg,marchetti2020mantra,marchetti2020multiple,deo2020trajectory,shafiee2021Introvert,mangalam2021ynet,yue2022nsf}. Recently, graph neural network-based methods, which define a graph representation with pedestrian nodes and interaction edges, are successfully incorporated into the trajectory prediction. Graph Convolutional Networks (GCNs)~\cite{kipf2016semi,mohamed2020social,sun2020rsbg,bae2021dmrgcn}, Graph Attention networks (GATs) \cite{velivckovic2018graph,huang2019stgat,kosaraju2019social,liang2020garden,liang2020simaug,Shi2021sgcn,bae2022npsn}, and transformers~\cite{yu2020spatio,yuan2021agent,gu2022mid,monti2022stt,bae2022gpgraph,wen2022socialode,wong2022v2net} are utilized to update the graph-structured pedestrian features.

\begin{figure}[t]
\centering
\includegraphics[width=\linewidth,trim={0 3mm 0 0},clip]{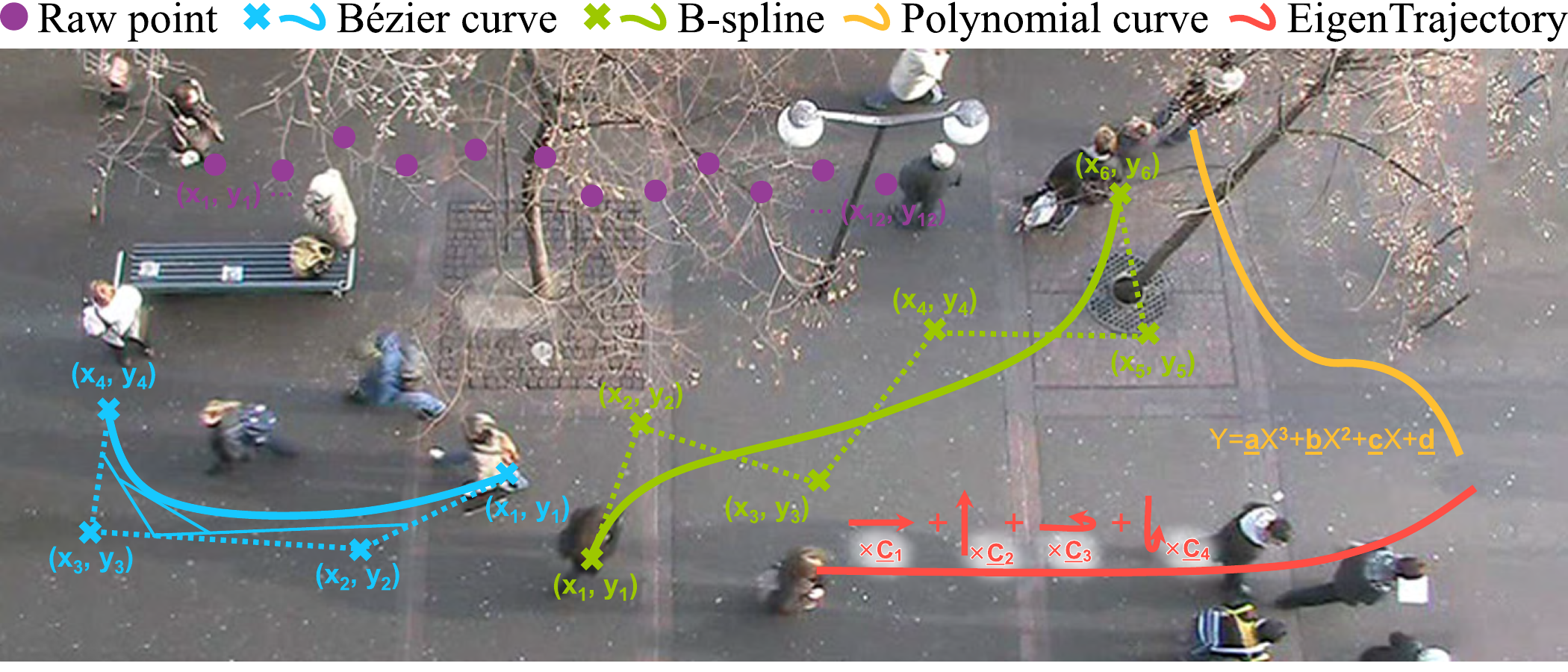}
\vspace{-5mm}
\caption{Visualization of the trajectory prediction strategies. The most widely used method is to directly predict 12 future raw points. The Bézier curve and B-spline are defined by a set of control points, and roughly model the movements. Our method can accurately represent pedestrian paths with a linear combination of our novel \ET~descriptors.}
\label{fig:curve}
\end{figure}

From the perspective of trajectory forecasting, probabilistic inferences are studied for multi-modal trajectory prediction. Gaussian modeling~\cite{alahi2016social,bae2021dmrgcn,mohamed2020social,shi2020multimodal,yu2020spatio,li2020Evolvegraph,shi2021socialdpf,yao2021bitrap,Shi2021sgcn,xu2022tgnn} estimates the bi-variate parameters of distribution to represent the possibility of moving routes. The Generative Adversarial Network (GAN)~\cite{gupta2018social,sadeghian2019sophie,kosaraju2019social,zhao2019matf,sun2020reciprocal,li2019idl,liang2021tpnms,dendorfer2021mggan,huang2019stgat} introduces additional discriminators to generate realistic paths. Conditional Variational AutoEncoder (CVAE)-based methods~\cite{lee2017desire,li2019conditional,ivanovic2019trajectron,bhattacharyya2020conditional,salzmann2020trajectron++,mangalam2020pecnet,chen2021disdis,sun2021pccsnet,lee2022musevae,wang2022stepwise,xu2022groupnet,xu2022socialvae} utilize random latent vectors as features for each person's walking direction and speed. They infer future trajectories recurrently \cite{alahi2016social,gupta2018social,bisagno2018group,pfeiffer2018,zhang2019srlstm,xu2020cflstm,salzmann2020trajectron++,ma2020autotrajectory,zhao2021experttraj,chen2021disdis,lee2022musevae,gu2022mid,marchetti2022smemo,navarro2022social} or simultaneously~\cite{mohamed2020social,bae2021dmrgcn,Shi2021sgcn,li2021stcnet,shi2022social,bae2023graphtern} in common. These approaches directly infer spatial coordinates in the Euclidean space without interpretable abstraction, which can suffer from an overfitting problem due to high dimensionality, and noisy paths can be generated as illustrated in~\cref{fig:curve}.

\subsection{Parametric Trajectory Descriptor}
To achieve a high-level abstraction of trajectory data, several recent works have introduced parametric trajectory descriptors. Hug \etal~\cite{hug2020bezier} propose a probabilistic, parametric $\mathcal{N}$-curve model based on the Bézier curve. To generate multi-modal predictions, multiple curves are computed from the Gaussian mixture model of random variables. Hug \etal~\cite{hug2022bezier} also introduce a variation of the Mixture Density Networks which is operated in the Bézier curve domain instead of the raw data. A Gaussian process-based refinement framework is adopted in their model. Alternatively, another work in~\cite{jazayeri2022bspline} represents each trajectory using a B-spline curve, and predicts the coefficients of this curve with the CVAE and inverse reinforcement learning.

The Bézier curve and B-spline use a Bernstein polynomial and piecewise polynomial as basis functions, respectively. These methods interpolate points on the curve through a weighted sum of a set of control points, but are still costly because it requires a large number of two-dimensional control points. One possible way to tackle this is to use polynomial curves; however, there are limitations on the precise path representations, and the abstract polynomial coefficients are difficult to learn. 
More fundamentally, although these works can successfully predict a temporally-smooth walking path using well-known curve functions, 
such parametric descriptors may be hard to fit irregular trajectories of human pedestrians.

\section{Methodology}
Our approach aims at learning an intuitive abstraction of human movements. To overcome the limited naturalness of parametric curves, we adopt a data-driven approach to leverage the distributions of human trajectories in the real-world data. Additionally, to address the dimensionality issue, we propose a low-rank approximation strategy to succinctly represent human trajectories in a more compact space than the Euclidean space. 
We achieve our goals by creating a novel trajectory descriptor specialized for representing human movements. 

We start with a definition of trajectory prediction in~\cref{sec:problem_definition}. We then introduce the mathematical formulations of \ET~descriptor and transformations between the Euclidean and \ET~spaces in~\cref{sec:math_formulation}. Using our descriptor, we describe how well existing trajectory forecasting baseline models can be operated in our novel space in~\cref{sec:forecasting_in_eigentrajectory}.

\subsection{Problem Definition}
\label{sec:problem_definition}
The problem of trajectory prediction involves forecasting the future paths of agents in an environment from their path histories. Here, a trajectory can be represented by a temporal series of spatial points. Formally, the observation trajectory with length $T_{obs}$ can be represented as $\bm{A}_n\!=\!\{ (x_n^t, y_n^t)\,|\,t\!\in\![1, ..., T_{obs}] \}$, where $(x_n^t, y_n^t)$ is the 2D spatial coordinate of a pedestrian $n$ at specific time $t$. Similarly, a ground truth future trajectory for prediction time length $T_{pred}$ can be defined as $\bm{B}_n\!=\!\{ (x_n^t, y_n^t)\,|\,t\!\in\![T_{obs}\!+\!1, ..., T_{pred}] \}$. 
A trajectory problem is defined as a pair $[\bm{A}, \bm{B}]$.
The goal of a forecasting model is to predict $s$ possible multi-modal future trajectories $\hat{\bm{B}}^s$ given observation $\bm{A}$ as input.

\subsection{EigenTrajectory (\ETtitle) Descriptor}
\label{sec:math_formulation}
Inspired by successful uses of Eigenvalue decomposition and Singular Value Decomposition (SVD) in extracting key feature information from raw data in various domains~\cite{turk1991eigenfaces,levin2007matting,levin2008matting,wang2018matting,shin2023ssgnet,jin2022eigenlanes,park2022eigencontours}, we propose an SVD-based approach for representing the gist of human trajectories. 

We assume that pedestrian movements are mainly represented by directions and speed variations that can be captured via principal components in the data. Since a little noise from people's staggering or tracking can also be added into the representation, we assume that using the top $k$ components in the eigenvectors would be sufficient to correspond to the raw trajectory data.
To fully benefit from the successful dimensionality reduction of SVD, we also apply it into our framework to represent spatio-temporal trajectory data as a linear combination of a set of eigenvectors.
The proposed \ET~descriptor consists of two elements: $k$ most representative singular vectors of trajectories and their combination coefficients that can explain any given trajectory.

\vspace{1mm}\noindent\textbf{Mathematical formulation.}\quad
To make an efficient trajectory descriptor, we need to find the principal components of the trajectory distribution. In mathematical terms, our goal is to obtain the eigenvectors of a covariance matrix from raw trajectory data. 
We first construct the trajectory matrices $\bm{A}$ and $\bm{B}$ by stacking all pedestrians' observations and predicted trajectories from a training set. We then apply SVD to both trajectory matrices $\bm{A}$ and $\bm{B}$ as follows:
\begin{equation}
   \bm{A} = \bm{U}_{\!obs} \bm{\mathit{\Sigma}}_{\!obs} \bm{V}^\top_{\!obs}, ~~~~~ \bm{B} = \bm{U}_{\!pred} \bm{\mathit{\Sigma}}_{\!pred} \bm{V}^\top_{\!pred},
    \label{eq:svd}
\end{equation}
where $\bm{U}\!\!=\![\bm{u}_1, \cdots, \bm{u}_L]$ and $\bm{V}\!\!=\![\bm{v}_1, \cdots, \bm{v}_N]$ are orthogonal matrices and $\bm{\mathit{\Sigma}}$ is a diagonal matrix, consisting of singular values $\sigma_1\!\geq\!\sigma_2\!\geq\!\cdots\!\geq\!\sigma_r\!>\!0$. Here, $L$ is the number of dimensions (\ie, the number of parameters) of the trajectory, $L=2\!\,\times\!\,T_{obs}$ for observation and $L=2\!\,\times\!\,(T_{pred}\!\,-\!\,T_{obs})$ for the predicted trajectory. $N$ is the number of pedestrians in the whole dataset, and $r$ is the rank of $\bm{A}$ and $\bm{B}$.

Since each trajectory can be more or less dominant to form a unique eigenvector, the resulting eigenvectors can be considered as a set of motion features that can jointly characterize trajectory variations.

\begin{figure}[t]
\centering
\includegraphics[width=\linewidth,trim={0 0 0 0},clip]{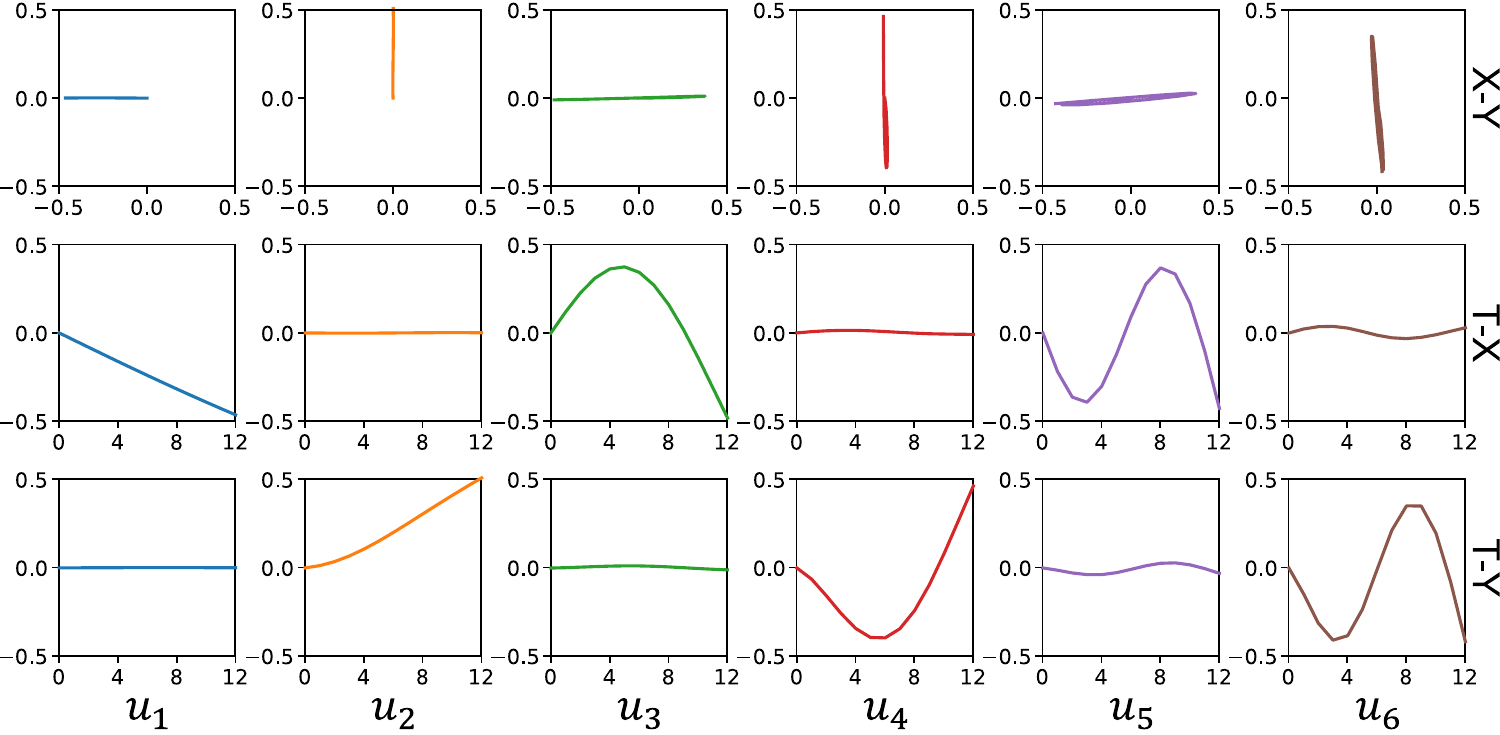}
\vspace{-5mm}
\caption{Visualization of the first six  \ET~descriptors $\bm{u}_1, \bm{u}_2, ..., \bm{u}_6$ learned from the ETH/UCY dataset. The top row shows the examples of the \ET~descriptors in 2D Euclidean space. The middle and bottom rows illustrate the temporal variations of the $x$ and $y$ positions, respectively.}
\vspace{-1mm}
\label{fig:eigenvector}
\end{figure}

\vspace{1mm}\noindent\textbf{Rank-$\bm{k}$ approximation.}\quad
Based on the assumption that human trajectories can be described using a few key parameters on directions and speed, we can approximate trajectory $\tilde{\bm{A}}_n$ using a linear combination of the first $k$ left singular vectors $\bm{U}_{k} = \bm{u}_1, \cdots, \bm{u}_k$ as:
\begin{equation}
    \tilde{\bm{A}}_n = \bm{U}_{\!obs,k} \,\bm{c}_{obs,n} = [\bm{u}_1, \cdots, \bm{u}_k]_{obs} \,\bm{c}_{obs,n},
    \label{eq:truncated_obs}
\end{equation}
where $\bm{c}_{obs,n}$ denotes a set of coefficients specifying the relevance of each principal component and 
$\tilde{\bm{A}}_n$ is known as the best rank-$k$ approximation of $\bm{A}$. 
Here, $k$ representative vectors are the eigenvectors of trajectories $\bm{A}\bm{A}^\top$ that is used to summarize the raw trajectories in our trajectory prediction approach. 
We apply the same approximation to the predicted trajectory $\bm{B}$ as follows:
\begin{equation}
   \tilde{\bm{B}}_n = \bm{U}_{\!pred,k} \,\bm{c}_{pred,n} = [\bm{u}_1, \cdots, \bm{u}_k]_{pred} \,\bm{c}_{pred,n}.
   \label{eq:truncated_pred}
\end{equation}
They can be regarded as the $k$ principal components. 
A pair $(\bm{c}_{n}, \bm{U}_{k})$ of the coefficients and rank-$k$ singular vectors constitutes the \ET~descriptor. 

In~\cref{fig:eigenvector}, we visualize the \ET~descriptors for predicted trajectory $\bm{B}$ and $k=6$ 
where the trajectories represented in rank-$k$ vectors explain the spatial displacement of pedestrians over time.
Specifically, $\bm{u}_1$ and $\bm{u}_2$ encode the constant velocity motion in the $x$ and $y$ directions, respectively. The reason why both eigenvectors appear first is that most people in the dataset walk at a constant speed while maintaining their directions. Next, $\bm{u}_3$ and $\bm{u}_4$ show velocity changes in the $x$ and $y$-axis directions, respectively. Using the combination of the vectors, it is feasible to learn complex movements, \eg, a person is going straight in the beginning, but then slows down or turns left/right. In $\bm{u}_5$ and $\bm{u}_6$, we also observe a velocity change, but \ET~descriptors can represent a more detailed trajectory by combining them with the previous eigenvector. We call the space spanned by \ET~descriptors as the \ET~space.

\begin{figure}[t]
\centering
\includegraphics[width=\columnwidth,trim={0 0 0 0},clip]{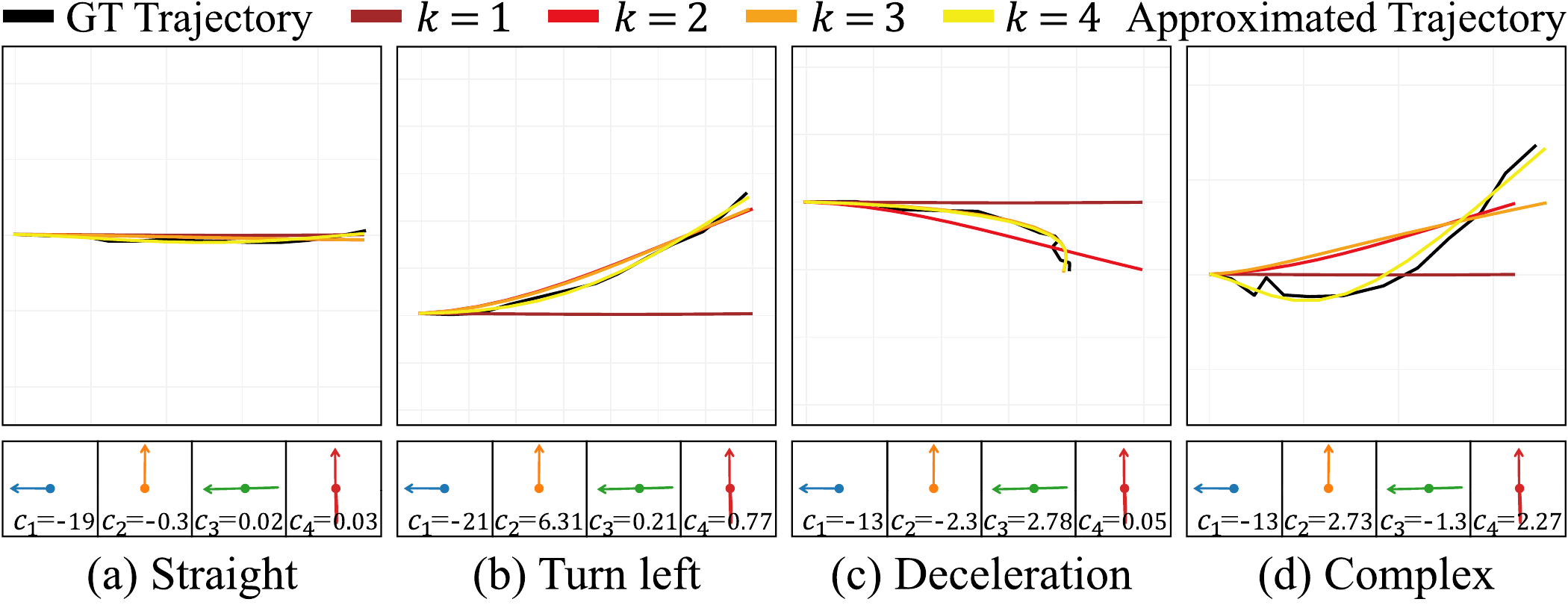}
\vspace{-5mm}
\caption{Visualization of the rank-$k$ trajectory approximation. Original trajectory sequences including straight, turning left, deceleration, and complex movement are successfully reconstructed by our \ET~descriptor when increasing the $k$ values. In this example, all four GT trajectories are sampled from the ZARA1 scene.}
\label{fig:eigenvector_change_k}
\end{figure}

\vspace{1mm}\noindent\textbf{Transformation between spaces.}\quad
Using the \ET~descriptors, we can project a trajectory defined in Euclidean space into the \ET~space. Given trajectories $\bm{A}_n$ and $\bm{B}_n$, we project it as:
\vspace{-1mm}
\begin{equation}
    \bm{c}_{obs,n} = \bm{U}_{obs,k}^\top \,\bm{A}_n, ~~~~~ \bm{c}_{pred,n} = \bm{U}_{pred,k}^\top \,\bm{B}_n,
    \label{eq:to_eigentrajectory}
\end{equation}
where $\bm{c}$ is a coefficient vector defined in the \ET~space. Each coefficient determines how much a corresponding \ET~descriptors affects a pedestrian path.

Obviously, the inverse transformation of~\cref{eq:to_eigentrajectory} is available. At this time, the reconstructed path is low-rank approximated. Since $\bm{U}$ is an orthogonal matrix, it can be invertible through a transpose operation as follows:
\begin{equation}
    \widetilde{\bm{A}}_n = \bm{U}_{obs,k} \,\bm{c}_{obs,n}, ~~~~~ \widetilde{\bm{B}}_n = \bm{U}_{pred,k} \,\bm{c}_{pred,n}.
    \label{eq:to_euclidean}
\end{equation}

As visualized in~\cref{fig:eigenvector_change_k}, a simple straight path can be expressed with only one \ET~descriptor ($k\!=\!1$). However, the more complex the trajectory, the larger value of $k$ is needed. Nevertheless, most paths can be represented just with $k$, which is small enough compared to the original trajectory dimension $k \le L$.

\subsection{Forecasting in the \ETtitle~Space}
\label{sec:forecasting_in_eigentrajectory}
Using the \ET~descriptor and transformation operators, we optimize each module designed for trajectory forecasting in our \ET~space. 
In addition, we propose a trajectory anchor to ensure the diversity of the predicted trajectories. This enables the off-the-shelf forecasting model to take the full benefits of the \ET~descriptor in an end-to-end manner.

\vspace{1mm}\noindent\textbf{Trajectory prediction using the \ETbold~descriptor.}\quad
Given a trajectory prediction problem $[\bm{A}, \bm{B}]$, using~\cref{eq:truncated_obs,eq:truncated_pred}, we use the training data to compute the \ET~descriptors for the observed and predicted trajectories, denoted by $(\bm{c}_{obs,n},\bm{U}_{\!obs,k})$ and $(\bm{c}_{pred,n}, \bm{U}_{\!pred,k})$, respectively.

\begin{figure}[t]
\centering
\includegraphics[width=\linewidth,trim={0 17mm 0 0},clip]{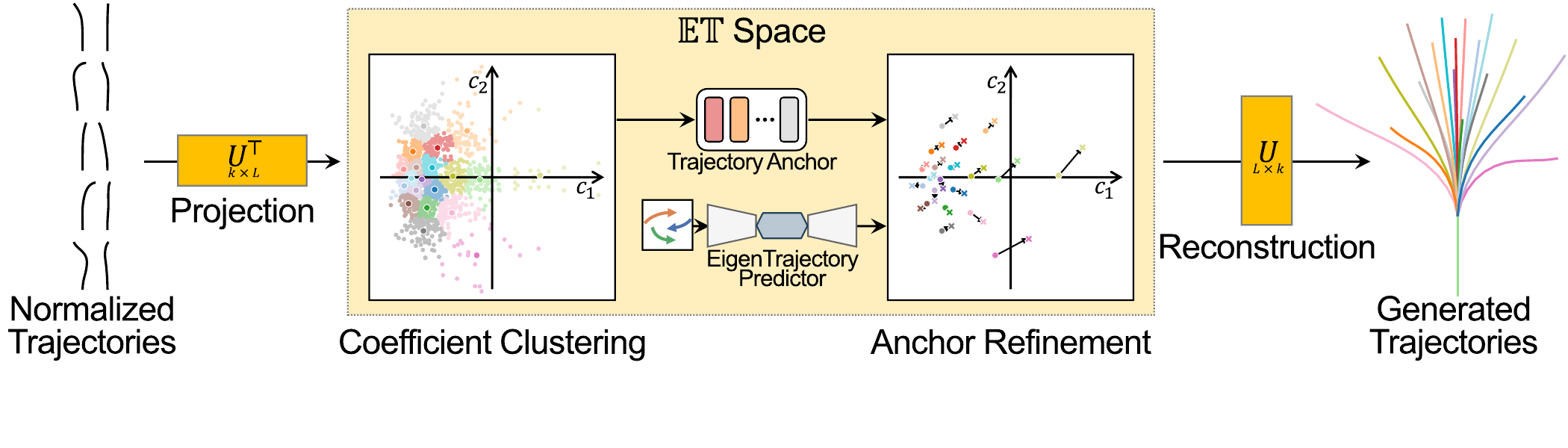}
\vspace{-5.5mm}
\caption{An overview of the trajectory anchor generation and anchor refinement step in our \ET~space.}
\vspace{-3mm}
\label{fig:anchor}
\end{figure}

\vspace{1mm}\noindent\textbf{Trajectory anchor generation.}\quad
Anchor-based methods are widely used in various fields, especially object detection~\cite{redmon2016yolo,liu2016ssd}. In the trajectory prediction, social anchors, offering the interpretability of egocentric characteristics, are introduced in~\cite{kothari2021interpretable,zhao2020tnt,xu2022remember,chai2019multipath}. However, these approaches mainly use hand-crafted anchors and work in Euclidean or Polar coordinate systems. In this paper, we introduce a data-driven anchor used in our novel \ET~space, which covers all feasible and diverse multi-modal futures.

We first normalize the translation, rotation, and speed of the trajectories in the training dataset as visualized in~\cref{fig:anchor}. Next, we project all normalized $N$ paths into the \ET~space using~\cref{eq:to_eigentrajectory}. We then cluster them in the \ET~space instead of using them in the original Euclidean space. Since there is a duality between the Euclidean space and the \ET~space with an isometry $\bm{U}_{pred,k}^\top$, the distance between the trajectories in the Euclidean space $\| \widetilde{\bm{B}}_i - \widetilde{\bm{B}}_j\|$ is equal to those between the corresponding coefficient vectors in the \ET~space $\| \bm{c}_{pred,i} - \bm{c}_{pred,j}\|$. Hence, the clustering can be performed to yield the same results in both spaces, but it can be done more reliably and more efficiently in the \ET~space. Note that the clustering becomes more difficult because of the well-known curse of dimensionality~\cite{dimensionality1,dimensionality2} and this process is executed once at the initialization step before training the model. We use these clustered $s$ centroids $\bar{c}_n^s$ as a trajectory coefficient anchor, and compute the correction offset for refinement in the prediction module.

\vspace{1mm}\noindent\textbf{Observations in the \ETbold~space.}\quad
As a next step, we project the input observed path into the \ET~space using the \ET~descriptor $U_{obs,k}$ calculated in~\cref{eq:truncated_obs} and transformation function in~\cref{eq:to_eigentrajectory}. This higher level of abstraction for the input data representations according to the change of the trajectory descriptor has a fundamental impact on the design of subsequent modules. In addition, the forecasting model can focus on the principal component of physical movements and has an additional effort of trajectory noise suppression by reducing the dimensionality.

\vspace{1mm}\noindent\textbf{History encoder in the \ETbold~space.}\quad
The set of \ET~descriptor coefficients for the observation is then fed to the trajectory forecasting model as input. Since the Euclidean coordinate sequence $\bm{A}_n$ can be transformed into a linear combination of \ET~descriptors $c_{obs,n}$, the model is able to omit a process for finding physical motion properties from the raw point and to immediately use the movement patterns.

\begin{figure}[t]
\centering
\includegraphics[width=\linewidth,trim={0 58mm 0 0},clip]{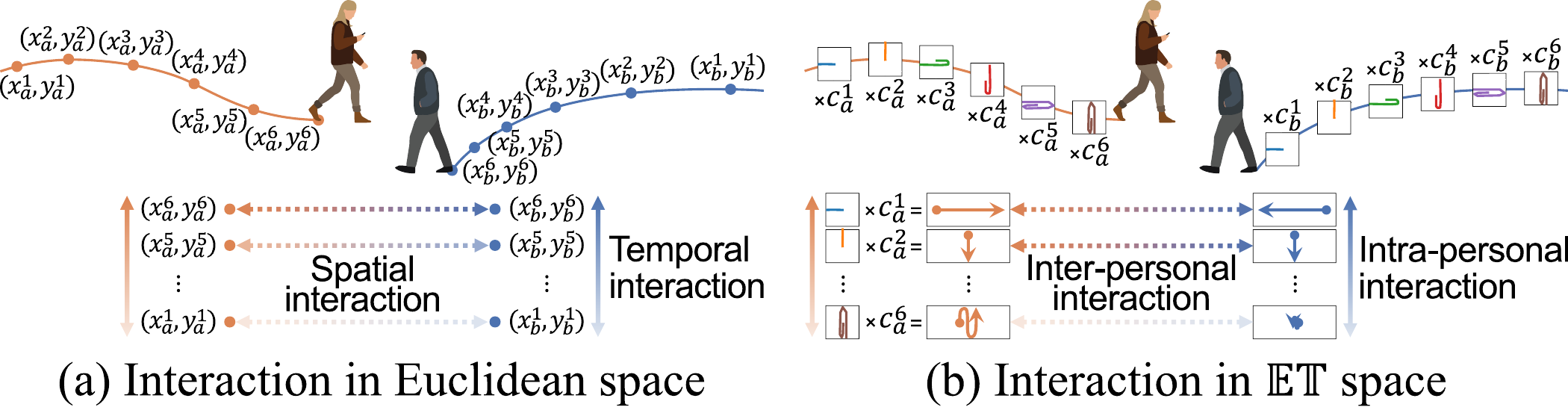}
\vspace{-5mm}
\caption{Illustration of the social interaction modeling strategies. (a) Social interactions between pedestrians are captured and gathered for each footstep in the Euclidean space. (b) The intra- and inter-personal interactions are modeled by \ET~descriptor coefficients in our \ET~space.}
\label{fig:interaction}
\vspace{-0.5mm}
\end{figure}

\vspace{1mm}\noindent\textbf{Interaction modeling in the \ETbold~space.}\quad
Since the method for the trajectory description uses the \ET~space, it is necessary to define social interactions in the same space. The recent models try to design interaction modules through a separation of the social relations between agents into spatial and temporal perspectives~\cite{mohamed2020social,Shi2021sgcn,yu2020spatio}. We map these spatial and temporal interactions one-to-one. To do this, we extract both the inter-personal and intra-personal interactions from our \ET~space, as illustrated in~\cref{fig:interaction}. In the inter-personal point of view, the relation between agents is captured with respect to the coefficients. Here, people with similar coefficients may be peers walking in a same direction; otherwise, they may need to avoid collisions. This information from each coefficient dimension is then aggregated in the intra-personal interaction stage to capture a rich social relation feature. We directly utilize the well-designed GNN and attention mechanism of the existing off-the-shelf models in our system with only a minor change for the dimension.

\vspace{1mm}\noindent\textbf{Extrapolation decoder in the \ETbold~space.}\quad
With the rich features in the \ET~space, we also predict future trajectories in the same space. To fully reflect the multi-modal and non-linear characteristics of human motions, we introduce the trajectory anchor-based method defined in the preparation stage. Using the trajectory anchor, the model predicts the correction vector $f$, which has the same shape as the anchor. This correction vector is added to the initial value of the trajectory anchor to obtain the final $s$ trajectory coefficients as follows:
\begin{equation}
    \hat{c}_n^s = \bar{c}_n^s + f_n^s.
\end{equation}

\vspace{1mm}\noindent\textbf{Reconstruction in the Euclidean space.}\quad
As a last step, we reconstruct the $s$ refined coefficients $\hat{c}_{pred,n}^s$, and convert them into the full trajectory points. For a fair comparison with the existing methods, we transform the model output to the original Euclidean coordinate system using~\cref{eq:to_euclidean}. Note that in the inference stage, we utilize $\bm{U}_{obs}$ and $\bm{U}_{pred}$ obtained from the training datasets for the transformation and reconstruction. The reconstructed trajectories are then used for loss functions and evaluation metrics.

\begin{table}[t]
\large
\centering
\resizebox{\linewidth}{!}{
\begin{tabular}{cccccccc}
\toprule
Descriptor                                      & Dim & ETH      & HOTEL   & UNIV    & ZARA1   & ZARA2   & AVG     \\ \midrule
Linear interpolation                            & 4   & \!386\,/\,703\! & \!143\,/\,215\! & \!163\,/\,344\! & \!256\,/\,426\! & \!155\,/\,232\! & \!221\,/\,384\! \\ 
Bézier curve~\cite{hug2020bezier,hug2022bezier} & 12  & 24 / 47  & 17 / 23 & 18 / 20 & 18 / 19 & 18 / 18 & 19 / 25 \\ 
B-spline~\cite{jazayeri2022bspline}             & 12  & 25 / 45  & 19 / 24 & 21 / 21 & 20 / 20 & 21 / 20 & 21 / 26 \\ \cmidrule(lr){1-8}
\multirow{5}{*}{\ETbold~\tbf{descriptor}}       & 4   & 40 / 107 & 20 / 36 & 14 / 45 & 13 / 38 & 08 / 28 & 19 / 51 \\
                                                & 6   & 27 / 65  & 14 / 27 & 08 / 23 & 07 / 20 & 04 / 14 & 12 / 30 \\
                                                & 8   & 19 / 50  & 11 / 23 & 05 / 15 & 04 / 12 & 03 / 09 & 08 / 22 \\
                                                & 10  & \tul{13} / \tul{39}  & \tul{07} / \tul{20} & \tul{03} / \tul{11} & \tul{03} / \tul{09} & \tul{02} / \tul{06} & \tul{06} / \tul{17} \\
                                                & 12  & \tbf{06} / \tbf{33}  & \tbf{03} / \tbf{17} & \tbf{01} / \tbf{08} & \tbf{01} / \tbf{07} & \tbf{01} / \tbf{05} & \tbf{03} / \tbf{14} \\
\bottomrule
\end{tabular}
}
\vspace{-1mm}
\caption{Comparison of the trajectory approximation accuracy of each descriptor. Both the observation and the prediction are curves fitted or approximated to each descriptor, reconstructed in the Euclidean space to fairly measure the error. The averaged $L_2$ distances between ground truth and reconstructed trajectory points are reported (Observation/prediction, Dim: Dimension of descriptor, Unit: $mm$). \tbf{Bold}: Best, \tul{underline}: Second best.}
\label{tab:trajectory_recon_error}
\end{table}

\newcommand{\ADE}{ADE\,\raisebox{0.2ex}{\darrow}}
\newcommand{\FDE}{FDE\,\raisebox{0.2ex}{\darrow}}
\newcommand{\COL}{COL\,\raisebox{0.2ex}{\darrow}}
\newcommand{\TCC}{TCC\,\raisebox{0.2ex}{\uarrow}}
\newcommand{\GAI}{\kern0.5ex Gain\,\raisebox{0.2ex}{\uarrow}\kern0.5ex}

\begin{table*}[t]
\large
\centering
\vspace{1mm}
\begin{tabular}{@{}c@{}}
\resizebox{\linewidth}{!}{
\begin{tabular}{c c cccccccccc c c ccccccccc c}
\toprule
\kern4em & ~ & \multicolumn{4}{c}{STGCNN~\cite{mohamed2020social}} & ~ & \multicolumn{5}{c}{\tbf{EigenTrajectory\,-\,STGCNN}} & ~ & ~ & \multicolumn{4}{c}{SGCN~\cite{Shi2021sgcn}} & ~~ & \multicolumn{5}{c}{\tbf{EigenTrajectory\,-\,SGCN}} \\ \cmidrule{3-6} \cmidrule(r){8-12} \cmidrule{15-18} \cmidrule(r){20-24}
      & & \ADE & \FDE & \TCC & \COL & & \ADE & \FDE & \TCC & \COL & \GAI & & & \ADE & \FDE & \TCC & \COL & & \ADE & \FDE & \TCC & \COL & \GAI \\ \midrule
ETH   & & 0.650 & 1.097 & \tbf{0.510} & 1.804 & & \tbf{0.365} & \tbf{0.582} & 0.471 & \tbf{1.050} & 46.9\% & & & 0.567 & 0.997 & \tbf{0.545} & 1.686 & & \tbf{0.360} & \tbf{0.565} & 0.438 & \tbf{1.160} & 43.3\% \\
HOTEL & & 0.496 & 0.858 & 0.270 & 3.936 & & \tbf{0.147} & \tbf{0.220} & \tbf{0.272} & \tbf{1.700} & 74.3\% & & & 0.308 & 0.533 & \tbf{0.295} & 2.523 & & \tbf{0.131} & \tbf{0.210} & 0.269 & \tbf{1.752} & 60.7\% \\
UNIV  & & 0.441 & 0.798 & 0.637 & 9.691 & & \tbf{0.246} & \tbf{0.427} & \tbf{0.751} & \tbf{8.548} & 46.5\% & & & 0.374 & 0.668 & 0.689 & \tbf{6.846} & & \tbf{0.244} & \tbf{0.428} & \tbf{0.791} & 8.362 & 35.9\% \\
ZARA1 & & 0.341 & 0.532 & 0.710 & 2.528 & & \tbf{0.217} & \tbf{0.393} & \tbf{0.808} & \tbf{1.396} & 26.1\% & & & 0.285 & 0.508 & 0.746 & \tbf{0.791} & & \tbf{0.200} & \tbf{0.347} & \tbf{0.841} & 1.147 & 31.6\% \\
ZARA2 & & 0.305 & 0.482 & 0.394 & 7.150 & & \tbf{0.168} & \tbf{0.290} & \tbf{0.645} & \tbf{6.212} & 39.7\% & & & 0.225 & 0.422 & 0.491 & \tbf{2.234} & & \tbf{0.153} & \tbf{0.261} & \tbf{0.611} & 5.892 & 38.1\% \\ \cmidrule(lr){1-24}
AVG   & & 0.447 & 0.753 & 0.504 & 5.022 & & \tbf{0.229} & \tbf{0.383} & \tbf{0.589} & \tbf{3.781} & 49.2\% & & & 0.352 & 0.626 & 0.553 & \tbf{2.816} & & \tbf{0.218} & \tbf{0.362} & \tbf{0.590} & 3.663 & 42.1\% \\ 
SDD   & & 20.76 & 33.18 & 0.471 & 0.679 & & \tbf{8.11}  & \tbf{13.35} & \tbf{0.578} & \tbf{0.541} & 59.8\% & & & 11.42 & 18.96 & 0.570 & 4.450 & & \tbf{8.05}  & \tbf{13.25} & \tbf{0.582} & \tbf{0.541} & 30.1\% \\
GCS   & & 14.72 & 23.87 & 0.698 & 3.921 & & \tbf{8.45}  & \tbf{14.49} & \tbf{0.834} & \tbf{3.354} & 39.3\% & & & 11.18 & 20.65 & 0.777 & \tbf{1.450} & & \tbf{7.86}  & \tbf{13.38} & \tbf{0.849} & 3.145 & 35.2\% \\ \bottomrule
\end{tabular}
} \\ \vspace{-4mm} \\
\resizebox{\linewidth}{!}{
\begin{tabular}{c c cccccccccc c c ccccccccc c}
\toprule
\kern4em & ~ & \multicolumn{4}{c}{PECNet~\cite{mangalam2020pecnet}} & ~ & \multicolumn{5}{c}{\tbf{EigenTrajectory\,-\,PECNet}} & ~ & ~ & \multicolumn{4}{c}{AgentFormer~\cite{yuan2021agent}} & ~~ & \multicolumn{5}{c}{\tbf{EigenTrajectory\,-\,AgentFormer}} \\ \cmidrule{3-6} \cmidrule(r){8-12} \cmidrule{15-18} \cmidrule(r){20-24}
      & & \ADE & \FDE & \TCC & \COL & & \ADE & \FDE & \TCC & \COL & \GAI & & & \ADE & \FDE & \TCC & \COL & & \ADE & \FDE & \TCC & \COL & \GAI \\ \midrule
ETH   & & 0.610 & 1.073 & \tbf{0.596} & 3.076 & & \tbf{0.365} & \tbf{0.572} & 0.580 & \tbf{1.215} & 46.7\% & & & 0.456 & 0.797 & \tbf{0.594} & 1.105 & & \tbf{0.362} & \tbf{0.568} & 0.487 & \tbf{1.105} & 28.7\%  \\
HOTEL & & 0.222 & 0.390 & \tbf{0.335} & 5.689 & & \tbf{0.132} & \tbf{0.211} & 0.298 & \tbf{1.192} & 45.9\% & & & \tbf{0.142} & 0.222 & \tbf{0.363} & \tbf{0.579} & & 0.147 & \tbf{0.222} & 0.267 & 1.866 & -0.1\%  \\
UNIV  & & 0.335 & 0.558 & 0.752 & \tbf{3.804} & & \tbf{0.244} & \tbf{0.432} & \tbf{0.765} & 8.310 & 22.6\% & & & 0.254 & 0.454 & \tbf{0.775} & \tbf{4.636} & & \tbf{0.244} & \tbf{0.430} & 0.747 & 8.416 & 5.4\%   \\
ZARA1 & & 0.250 & 0.448 & 0.808 & 2.993 & & \tbf{0.195} & \tbf{0.348} & \tbf{0.828} & \tbf{0.996} & 22.4\% & & & \tbf{0.176} & \tbf{0.303} & \tbf{0.839} & \tbf{0.235} & & 0.216 & 0.397 & 0.808 & 1.416 & \!-31.1\%\! \\
ZARA2 & & 0.186 & 0.332 & 0.596 & 4.910 & & \tbf{0.143} & \tbf{0.250} & \tbf{0.628} & \tbf{2.817} & 24.8\% & & & \tbf{0.141} & \tbf{0.237} & 0.565 & \tbf{1.186} & & 0.166 & 0.290 & \tbf{0.731} & 6.010 & \!-22.1\%\! \\ \cmidrule(lr){1-24}
AVG   & & 0.321 & 0.560 & 0.617 & 4.094 & & \tbf{0.216} & \tbf{0.362} & \tbf{0.620} & \tbf{2.906} & 35.3\% & & & 0.234 & 0.403 & \tbf{0.627} & \tbf{1.548} & & \tbf{0.227} & \tbf{0.381} & 0.608 & 3.763 & 5.3\%   \\ 
SDD   & & 9.97  & 15.89 & \tbf{0.647} & \tbf{1.444} & & \tbf{8.12}  & \tbf{13.10} & 0.575 & 2.970 & 17.6\% & & & 8.68  & 14.92 & \tbf{0.608} & \tbf{0.379} & & \tbf{8.10}  & \tbf{13.43} & 0.590 & 0.562 & 10.0\%  \\
GCS   & & 17.08 & 29.30 & 0.708 & \tbf{2.866} & & \tbf{7.42}  & \tbf{12.49} & \tbf{0.888} & 2.970 & 57.4\% & & & 10.18 & 16.91 & 0.840 & \tbf{2.319} & & \tbf{8.41}  & \tbf{14.56} & \tbf{0.889} & 3.263 & 13.9\%  \\ \bottomrule
\end{tabular}
}
\end{tabular}
\vspace{-1mm}
\caption{Comparison between EigenTrajectory and the Eucledian space for four state-of-the-art multi-modal trajectory prediction models, Social-STGCNN~\cite{mohamed2020social}, SGCN~\cite{Shi2021sgcn}, PECNet~\cite{mangalam2020pecnet} and AgentFormer~\cite{yuan2021agent}. The models are evaluated on the ETH~\cite{pellegrini2009you}, UCY~\cite{lerner2007crowdsbyexample}, SDD~\cite{robicquet2016learning} and GCS~\cite{yi2015understanding} datasets. Gain: performance improvement w.r.t FDE over the baseline models, Unit for ADE and FDE: meter, \tbf{Bold}: Best.}
\label{tab:eigentraj_trajresult}
\end{table*}

\subsection{Implementation Details}
We incorporate our \ET~into six state-of-the-art pedestrian trajectory forecasting baselines~\cite{mohamed2020social,Shi2021sgcn,mangalam2020pecnet,yuan2021agent,pang2021lbebm,mohamed2022socialimplicit}. To validate the generality of our \ET~space, we simply change the size of the input and output channel dimensions of them. We use a loss function which is a linear combination of three terms to train our EigenTrajectory models. First, we measure a Frobenius norm between the refined coefficients $c_{pred,n}$ to regress the ground truth coefficient $c_{pred}$. Here, the winner-takes-all process~\cite{rupprecht2017learning} is 
chosen to back-propagate only to the closest trajectory anchors for the training:
\begin{equation}
    \mathcal{L}_{\textit{coeff}} = \frac{1}{N} \sum_{n=1}^{N}\min_{i \in [1, ..., s]} \left\| \hat{c}_{pred,i}^s - c_{pred,i} \right\|.
\end{equation}

Next, we calculate a Euclidean norm between the reconstructed path and the ground-truth raw coordinate, and then average them along a time axis as shown below:
\begin{equation}
    \mathcal{L}_{\textit{dist}} = \frac{1}{N(T_{\!pred}\!-\!T_{\!obs})} \!\sum_{n=1}^{N} \!\sum_{\;t=T_{\!obs}\!+\!1}^{T_{\!pred}} \!\! \min_{i \in [1, ..., s]} \left\| \widehat{B}_{i,t}^s - B_{i,t} \right\|.
\end{equation}

Lastly, we impose an additional penalty for the last prediction frame. The ground-truth coefficient $c_{pred}$ obtained by the low-rank approximation tends to be more careless for the endpoint because it is designed to minimize the error of the whole path. We thus penalize the endpoint coordinate so that the model can learn to infer the correct destination:
\begin{equation}
    \mathcal{L}_{\textit{end}} = \frac{1}{N} \sum_{n=1}^{N}\min_{i \in [1, ..., s]} \left\| \widehat{B}_{i,T_{pred}}^s - B_{i,T_{pred}} \right\|.
\end{equation}

The final loss function is a linear combination of the three losses $\mathcal{L} = \mathcal{L}_{\textit{coeff}} + \alpha\mathcal{L}_{dist} + \beta\mathcal{L}_{\textit{end}}$. We empirically set $\alpha$ and $\beta$ to 1. 
We train our EigenTrajectory models with the AdamW optimizer~\cite{loshchilov2018decoupled} with a batch size of 128 and a learning rate of 0.001 for 256 epochs. The training time takes about a day on a machine with an NVIDIA 3090 GPU.

\section{Experiments}
In this section, we conduct comprehensive experiments on public benchmark datasets to verify the efficiency of our \ET~descriptor and the effectiveness of the \ET~space for the trajectory prediction. We first describe our experimental setup briefly in~\cref{sec:experiment_setup}. We then provide comparison results with other trajectory descriptors, various baselines, and state-of-the-art models in~\cref{sec:result}. Finally, we present the results of an extensive ablation study demonstrating the effect of each component of our method in \cref{sec:ablation}.

\subsection{Experimental Setup}
\label{sec:experiment_setup}
\vspace{0mm}\noindent\textbf{Datasets.}\quad
We conduct experiments on four public datasets: ETH~\cite{pellegrini2009you}, UCY~\cite{lerner2007crowdsbyexample}, Stanford Drone Dataset (SDD)~\cite{robicquet2016learning}, and the Grand Central Station (GCS)~\cite{yi2015understanding} datasets to compare our EigenTrajectory with state-of-the-art baselines and to check the performance improvement for trajectory forecasting. The ETH and UCY datasets consist of pedestrian trajectories across five unique scenes (ETH, Hotel, Univ, Zara1 and Zara2) with 1,536 pedestrians recorded in the world coordinates. Following previous works~\cite{alahi2016social,gupta2018social}, we adopt the standard leave-one-out strategy for the training and evaluation. SDD has 5,232 pedestrians in eight different university campus scenes at a top-down drone view. GCS shows a highly congested terminal scene with 12,684 pedestrians. We use a standard training and evaluation protocol~\cite{gupta2018social,huang2019stgat,mohamed2020social,Shi2021sgcn,bae2022npsn} in which the first 3.2 seconds ($T_{obs}\!=\!8$ frames) are observed, and succeeding 4.8 seconds ($T_{pred}\!-\!T_{obs}\!=\!12$ frames) are used for the trajectory prediction. 

\vspace{1mm}\noindent\textbf{Baseline models.}\quad
We evaluate an efficiency and a generality of our EigenTrajectory by incorporating it into the following state-of-the-art baseline models: Social-STGCNN~\cite{mohamed2020social}, SGCN~\cite{Shi2021sgcn}, PECNet~\cite{mangalam2020pecnet}, AgentFormer~\cite{yuan2021agent}, LB-EBM \cite{pang2021lbebm}, and Social-Implicit~\cite{mohamed2022socialimplicit}. For strictly fair comparison, we directly utilize authors' provided official source code for the vanilla baseline model designed in the Euclidean space. To validate the effectiveness of our \ET~space, we modify the input and output shape of the baseline models to take and predict the coefficients of the \ET~descriptor, instead of the direct use of the raw Euclidean coordinates like the conventional manner.

\begin{figure*}[t]
\centering
\includegraphics[width=\linewidth,trim={8mm 0 8mm 0},clip]{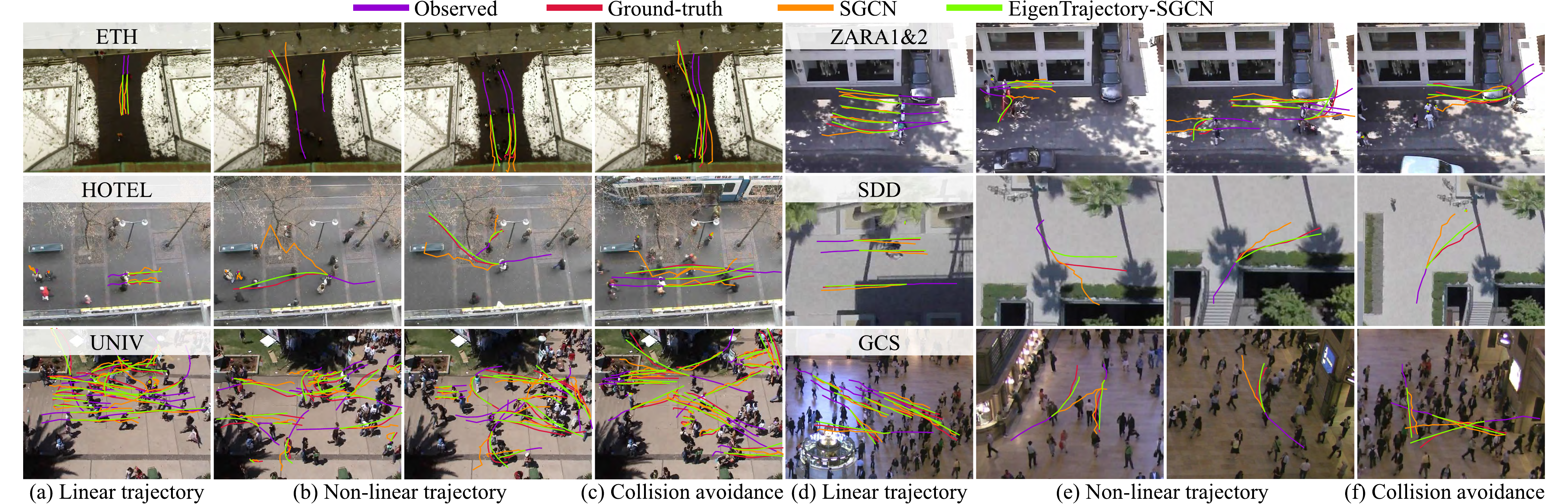}
\vspace{-5mm}
\caption{Examples of the prediction results from our EigenTrajectory predictor, compared to the baseline model in the conventional Euclidean space. To aid visualization,
the SGCN~\cite{Shi2021sgcn} is used, and we report one trajectory with the best FDE of $s\!=\!20$ samples each.}
\vspace{2mm}
\label{fig:qualitative}
\end{figure*}

\vspace{1mm}\noindent\textbf{Evaluation metrics.}\quad
To measure the prediction performance of the baseline models with our \ET~space, we use four metrics: Average Displacement Error (ADE), Final Displacement Error (FDE), Temporal Correlation Coefficient (TCC)~\cite{tao2020dynamic}, and Collision rate (COL)~\cite{liu2021snce}. The ADE and FDE compute the Euclidean distance between a predicted and a ground-truth trajectory and thier end-points, respectively. The TCC measures the Pearson correlation coefficient of motion patterns between a predicted and ground-truth trajectory, and the COL calculates the percentage of collision cases between agents on the predicted path. We use both the ADE and FDE as accuracy measures, and both the TCC and COL as reliability measures. Following~\cite{gupta2018social}, we generate $s=20$ samples, and then choose the best path to evaluate the multi-modal trajectory prediction performance.

\subsection{Evaluation Results}
\label{sec:result}
\vspace{0mm}\noindent\textbf{Evaluation of the trajectory descriptors.}\quad
First, we check the efficiency of our low-rank approximation method using the \ET~descriptor over other linear interpolation-based and parametric curve-based methods. The linear interpolation-based method predicts the coordinates of the first frame and the last frame of the predicted trajectory, and then equally divides and interpolates them to reconstruct the full frame points. Bézier curve and B-spline predict a set of control points and then compute the full coordinates through matrix multiplication with the Bernstein polynomial and piecewise polynomial basis, respectively. Following~\cite{hug2020bezier,hug2022bezier,jazayeri2022bspline}, the order of both curves is set to 5. 

\cref{tab:trajectory_recon_error} shows that our descriptor is 87\% more efficient than the results from the linear interpolation. This is because our \ET~descriptor can cover linear and non-linear movements of pedestrians with only 2-dimensional and 4-dimensional representations, respectively. Similarly, the \ET~descriptor shows superior low-rank approximation results even with the same $k=12$ dimension as both the Bézier curve and B-spline. We choose $k=6$ in our space design which shows similar or better approximation accuracy than that of the existing parametric curve.

\vspace{1mm}\noindent\textbf{Evaluation of the trajectory prediction.}\quad
Next, we evaluate the prediction space with the public trajectory prediction benchmarks. As reported in~\cref{tab:eigentraj_trajresult}, our EigenTrajectory framework achieves consistent performance improvements with all the baseline models. In particular, the prediction accuracy improves significantly, up to 74.3\%. These results demonstrate that the trajectory forecasting models can benefit from our \ET~space rather than dealing with raw data, and make the prediction task tractable when using trajectory anchors. In addition, most of the trajectory reliability metrics also achieve improvements by predicting accurate and stable movement patterns, as visualized in~\cref{fig:qualitative}. 

However, there are some limitations with our EigenTrajectory. The TCC metric is slightly worse in the ETH and hotel scenes because we remove the noisy motions through the truncation, but there are a lot of wobbling people in both scenes. In addition, due to the low-rank approximation, the macroscopic movements are leveraged, but the microscopic movements are removed. This is why there are more collisions in very complex UNIV and GCS scenes. In this case, it is important to find a good trade-off between the accuracy and reliability by adjusting $k$.

\vspace{1mm}\noindent\textbf{Comparison with the state-of-the-art models.}\quad
We compare our EigenTrajectory models with the state-of-the-art models. As shown in~\cref{tab:eigentraj_sota}, our EigenTrajectory achieves a better accuracy by taking full advantage of our data-driven trajectory descriptor than those of the previous models. Furthermore, our EigenTrajectory shows a higher improvement than other generalized approaches such as NCE~\cite{liu2021snce}, Causal~\cite{liu2021causal}, GP-Graph~\cite{bae2022gpgraph}, and NPSN~\cite{bae2022npsn}. While all of those approaches adhere to the conventional Euclidean space for the input and output, the introduction of our efficient descriptor can significantly improve the network ability.

\subsection{Ablation Studies}
\label{sec:ablation}

\vspace{0mm}\noindent\textbf{Effectiveness of each component.}\quad
We validate the effectiveness of each module optimized in our \ET~space. In~\cref{tab:ablation}, we only replace the input of the baseline models with the \ET~space and keep the output space as it is. In this case, we observe the marginal improvements. Next, we replace both the input and output space with the \ET~space, and achieve significant performance improvements. Here, we observe that the existing baseline network is struggling to predict all the points of the future path. Lastly, the best performance is obtained when our trajectory anchor is incorporated into the baseline model.

\vspace{1mm}\noindent\textbf{Trajectory anchors.}\quad
Next, we demonstrate the effectiveness of our \ET~space for trajectory anchor generation. \cref{tab:ablation} presents the results of an ablation study on the clustering space to obtain the data-driven trajectory anchors. The accuracy of the anchors from the \ET~space is much better compared to the Euclidean space. Additionally, the results are better than those of most state-of-the-art works, only with anchors without any refinement method. We think that the trajectory anchors offer an efficient initial trajectory candidate from the initial data, and better anchors can be estimated when clustering with lower dimensions.

\renewcommand{\dag}{$^\dagger$}

\newcommand{\psp}{\kern0.2ex}
\newcommand{\pslp}{\psp/\psp}
\newcommand{\nsp}{\kern-0.1ex}

\newcommand{\ETfull}{E\nsp i\nsp g\nsp e\nsp n\nsp\nsp\nsp T\nsp\nsp\nsp r\nsp a\nsp j\nsp e\nsp c\nsp t\nsp o\nsp r\nsp y}
\newcommand{\ETSTGCNN}{\ETfull\nsp -\nsp S\nsp T\nsp G\nsp C\nsp N\nsp N}
\newcommand{\ETSGCN}{\ETfull\nsp -\nsp S\nsp G\nsp C\nsp N}
\newcommand{\ETPECNet}{\ETfull\nsp -\nsp P\nsp E\nsp C\nsp N\nsp e\nsp t}
\newcommand{\ETAgentFormer}{\ETfull\nsp -\nsp A\nsp\nsp\nsp g\nsp e\nsp n\nsp t\nsp\nsp F\nsp\nsp\nsp o\nsp r\nsp m\nsp e\nsp r}
\newcommand{\ETLBEBM}{\ETfull\nsp -\nsp L\nsp B\nsp -\nsp E\nsp B\nsp M}
\newcommand{\ETDMRGCN}{\ETfull\nsp -\nsp D\nsp M\nsp R\nsp G\nsp C\nsp N}
\newcommand{\ETGraphTERN}{\ETfull\nsp -\nsp G\nsp\nsp r\nsp a\nsp p\nsp h\nsp -\nsp\nsp\nsp T\nsp\nsp E\nsp\nsp R\nsp\nsp N}
\newcommand{\ETImplicit}{\ETfull\nsp -\nsp I\nsp m\nsp p\nsp l\nsp i\nsp c\nsp i\nsp t}

\newcommand{\ETMID}{\ETfull\nsp -\nsp M\nsp I\nsp D}

\begin{table}[t]
    \Large
    \centering
    \resizebox{\linewidth}{!}{
\begin{tabular}{ccccccc}
\toprule
Model                                            & ETH & HOTEL & UNIV & ZARA1 & ZARA2 & AVG \\ \midrule
Social-LSTM~\cite{alahi2016social}               & 1.09\pslp2.35 & 0.79\pslp1.76 & 0.67\pslp1.40 & 0.47\pslp1.00 & 0.56\pslp1.17 & 0.72\pslp1.54 \\
Social-GAN~\cite{gupta2018social}                & 0.87\pslp1.62 & 0.67\pslp1.37 & 0.76\pslp1.52 & 0.35\pslp0.68 & 0.42\pslp0.84 & 0.61\pslp1.21 \\
STGAT~\cite{huang2019stgat}                      & 0.65\pslp1.12 & 0.35\pslp0.66 & 0.52\pslp1.10 & 0.34\pslp0.69 & 0.29\pslp0.60 & 0.43\pslp0.83 \\
Social-STGCNN~\cite{mohamed2020social}           & 0.64\pslp1.11 & 0.49\pslp0.85 & 0.44\pslp0.79 & 0.34\pslp0.53 & 0.30\pslp0.48 & 0.44\pslp0.75 \\
PECNet\dag~\cite{mangalam2020pecnet}             & 0.61\pslp1.07 & 0.22\pslp0.39 & 0.34\pslp0.56 & 0.25\pslp0.45 & 0.19\pslp0.33 & 0.32\pslp0.56 \\
Trajectron++\dag~\cite{salzmann2020trajectron++} & 0.61\pslp1.03 & 0.20\pslp0.28 & 0.30\pslp0.55 & 0.24\pslp0.41 & 0.18\pslp0.32 & 0.31\pslp0.52 \\
SGCN~\cite{Shi2021sgcn}                          & 0.57\pslp1.00 & 0.31\pslp0.53 & 0.37\pslp0.67 & 0.29\pslp0.51 & 0.22\pslp0.42 & 0.35\pslp0.63 \\
LB-EBM\dag~\cite{pang2021lbebm}                  & 0.60\pslp1.06 & 0.21\pslp0.38 & 0.28\pslp0.54 & 0.21\pslp0.39 & 0.15\pslp0.30 & 0.29\pslp0.53 \\
AgentFormer\dag~\cite{yuan2021agent}             & 0.46\pslp0.80 & 0.14\pslp0.22 & 0.25\pslp0.45 & 0.18\pslp\tbf{0.30} & 0.14\pslp\tbf{0.24} & 0.23\pslp0.40 \\
ExpertTraj~\cite{zhao2021experttraj}             & \tul{0.37}\pslp0.65 & \tbf{0.11}\pslp\tbf{0.15} & \tbf{0.20}\pslp\tul{0.44} & \tbf{0.15}\pslp\tul{0.31} & \tbf{0.12}\pslp0.26 & \tbf{0.19}\pslp0.36 \\
MemoNet~\cite{xu2022remember}                    & 0.40\pslp0.61 & \tbf{0.11}\pslp\tul{0.17} & 0.24\pslp\tbf{0.43} & 0.18\pslp0.32 & 0.14\pslp\tbf{0.24} & \tul{0.21}\pslp\tul{0.35} \\
Social-Implicit~\cite{mohamed2022socialimplicit} & 0.66\pslp1.44 & 0.20\pslp0.36 & 0.31\pslp0.60 & 0.25\pslp0.50 & 0.22\pslp0.43 & 0.33\pslp0.67 \\
MID~\cite{gu2022mid}                             & 0.39\pslp0.66 & 0.13\pslp0.22 & \tul{0.22}\pslp0.45 & \tul{0.17}\pslp\tbf{0.30} & \tul{0.13}\pslp0.27 & \tul{0.21}\pslp0.38 \\ \cmidrule(lr){1-7}

NCE-STGCNN~\cite{liu2021snce}                    & 0.67\pslp1.22 & 0.44\pslp0.68 & 0.47\pslp0.88 & 0.33\pslp0.52 & 0.29\pslp0.48 & 0.44\pslp0.76 \\
Causal-STGCNN~\cite{liu2021causal}               & 0.64\pslp1.00 & 0.38\pslp0.45 & 0.49\pslp0.81 & 0.34\pslp0.53 & 0.32\pslp0.49 & 0.43\pslp0.66 \\
GP\nsp -\nsp Graph-STGCNN~\cite{bae2022gpgraph}  & 0.48\pslp0.77 & 0.24\pslp0.40 & 0.29\pslp0.47 & 0.24\pslp0.40 & 0.23\pslp0.40 & 0.29\pslp0.49 \\
NPSN-STGCNN~\cite{bae2022npsn}                   & 0.44\pslp0.65 & 0.21\pslp0.34 & 0.28\pslp0.44 & 0.25\pslp0.43 & 0.22\pslp0.38 & 0.28\pslp0.45 \\ \cmidrule(lr){1-7}

\tbf{\ETSTGCNN}                                  & \tbf{0.36}\pslp0.58 & 0.15\pslp0.22 & 0.25\pslp\tbf{0.43} & 0.22\pslp0.39 & 0.17\pslp0.29 & 0.23\pslp0.38 \\
\tbf{\!\!\ETAgentFormer\!\!}                     & \tbf{0.36}\pslp\tul{0.57} & 0.15\pslp0.22 & 0.24\pslp\tbf{0.43} & 0.22\pslp0.40 & 0.17\pslp0.29 & 0.23\pslp0.38 \\
\tbf{\ETImplicit}                                & \tbf{0.36}\pslp\tul{0.57} & 0.13\pslp0.21 & 0.24\pslp\tbf{0.43} & 0.21\pslp0.37 & 0.15\pslp0.26 & 0.22\pslp0.37 \\
\tbf{\ETSGCN}                                    & \tbf{0.36}\pslp\tul{0.57} & 0.13\pslp0.21 & 0.24\pslp\tbf{0.43} & 0.20\pslp0.35 & 0.15\pslp0.26 & 0.22\pslp0.36 \\
\tbf{\ETPECNet}                                  & \tbf{0.36}\pslp\tul{0.57} & 0.13\pslp0.21 & 0.24\pslp\tbf{0.43} & 0.19\pslp0.35 & 0.14\pslp\tul{0.25} & 0.22\pslp0.36 \\
\tbf{\ETLBEBM}                                   & \tbf{0.36}\pslp\tbf{0.53} & \tul{0.12}\pslp0.19 & 0.24\pslp\tbf{0.43} & 0.19\pslp0.33 & 0.14\pslp\tbf{0.24} & \tul{0.21}\pslp\tbf{0.34} \\
\bottomrule
\end{tabular}
}
\vspace{-1mm}
\caption{Comparison EigenTrajectory methods with other state-of-the-art stochastic models (ADE/FDE, Unit: meter). {\begin{footnotesize}$\dagger$\end{footnotesize}}: Issues raised in the authors' GitHubs are fixed, \textbf{Bold}: Best, \underline{Underline}: 2$^\text{nd}$ Best.}
\label{tab:eigentraj_sota}
\end{table}

\begin{table}[t]
    \Large
    \centering
    \resizebox{\linewidth}{!}{
\begin{tabular}{ccccccc}
\toprule
Model & ETH & HOTEL & UNIV & ZARA1 & ZARA2 & AVG \\ \midrule

Baseline                     & 0.57 / 1.00 & 0.31 / 0.53 & 0.37 / 0.67 & 0.29 / 0.51 & 0.23 / 0.42 & 0.35 / 0.63 \\ \cmidrule(lr){1-7}
+ Observed trajectory        & 0.50 / 0.79 & 0.21 / 0.34 & \tul{0.33} / 0.62 & 0.27 / 0.48 & 0.26 / 0.44 & 0.31 / 0.53 \\
+ Predicted trajectory       & \tul{0.48} / \tul{0.70} & \tul{0.17} / \tul{0.27} & \tul{0.33} / \tul{0.61} & \tul{0.26} / \tul{0.47} & \tul{0.22} / \tul{0.40} & \tul{0.29} / \tul{0.49} \\
+ Trajectory anchor          & \tbf{0.36} / \tbf{0.57} & \tbf{0.13} / \tbf{0.21} & \tbf{0.24} / \tbf{0.43} & \tbf{0.20} / \tbf{0.35} & \tbf{0.15} / \tbf{0.26} & \tbf{0.22} / \tbf{0.36} \\ \cmidrule(lr){1-7}
Euclidean anchor             & 0.47 / 0.76 & 0.18 / 0.30 & 0.36 / 0.69 & 0.35 / 0.67 & 0.25 / 0.48 & 0.32 / 0.58 \\ 
\ET~anchor & \tbf{0.36} / \tbf{0.57} & \tbf{0.14} / \tbf{0.23} & \tbf{0.24} / \tbf{0.43} & \tbf{0.22} / \tbf{0.40} & \tbf{0.17} / \tbf{0.29} & \tbf{0.23} / \tbf{0.38} \\ 

\bottomrule
\end{tabular}
}
\vspace{-1mm}
\caption{Ablation study on trajectory anchor generated in different spaces on SGCN~\cite{Shi2021sgcn} (ADE/FDE, Unit: meter). \textbf{Bold}: Best.}
\label{tab:ablation}
\end{table}

\vspace{1mm}\noindent\textbf{Non-linear trajectories.}\quad
We evaluate how well the \ET~descriptor represent and predict non-linear trajectories. 
Following~\cite{gupta2018social}, we evaluate a case in which the linear approximation error of the future path is more than $0.02m$.
In~\cref{tab:ablation_nonlinear}, our \ET~space has a smaller performance drop, compared to the conventional Euclidean space. The prediction ability of a linear path is already saturated, so it is important to design a model to handle non-linear paths well. Since most of the paths in the dataset are straight, the output trajectories tend to be smooth during training. Nevertheless, our \ET~space shows a robust performance for these non-linear trajectories because our \ET~descriptor can explicitly represent non-linearity with the combinations of \ET~coefficients.

\vspace{1mm}\noindent\textbf{Trajectory perturbation.}\quad
Lastly, we examine the robustness of the prediction to input noise. To this end, we measure the change in prediction accuracy after adding a little Gaussian noise. In~\cref{tab:ablation_noise}, our EigenTrajectory has a marginal performance drop when noise exists, compared to the baselines. In our \ET~space, because the principal motion pattern components are only left through the rank-$k$ approximation, the negative effect of noise can be mitigated.

\begin{table}[t]
    \Large
    \centering
    \resizebox{\linewidth}{!}{
\begin{tabular}{c c cccccc}
\toprule
Space & Type & ETH & HOTEL & UNIV & ZARA1 & ZARA2 & AVG \\ \midrule

\multirow{2}{*}{Euclidean} & All & 0.57 / 1.00 & 0.31 / 0.53 & 0.37 / 0.67 & 0.29 / 0.51 & 0.23 / 0.42 & 0.35 / 0.63 \\
                           & NL  & 0.65 / 1.16 & 0.46 / 0.82 & 0.49 / 0.92 & 0.39 / 0.78 & 0.50 / 1.03 & 0.50 / 0.94 \\ \cmidrule(lr){1-8}
\multicolumn{2}{c}{Diff.} & \!-0.08\,/-0.17 & \!-0.15\,/-0.29 & \!-0.12\,/-0.25 & \!-0.11\,/-0.28 & \!-0.27\,/-0.61 & \!-0.15\,/-0.32 \\ \midrule
\multirow{2}{*}{\tbf{~~~\ETbold}} & All & 0.36 / 0.57 & 0.13 / 0.21 & 0.24 / 0.43 & 0.20 / 0.35 & 0.15 / 0.26 & 0.22 / 0.36 \\
                                  & NL  & 0.41 / 0.65 & 0.19 / 0.31 & 0.32 / 0.55 & 0.25 / 0.43 & 0.32 / 0.53 & 0.30 / 0.49 \\ \cmidrule(lr){1-8}
\multicolumn{2}{c}{Diff.} & \!\tbf{-0.05}\,/\tbf{-0.08} & \!\tbf{-0.06}\,/\tbf{-0.10} & \!\tbf{-0.08}\,/\tbf{-0.12} & \!\tbf{-0.05}\,/\tbf{-0.08} & \!\tbf{-0.17}\,/\tbf{-0.27} & \!\tbf{-0.08}\,/\tbf{-0.13} \\
\bottomrule
\end{tabular}
}
\vspace{-1mm}
\caption{Ablation study on non-linear trajectories on SGCN~\cite{Shi2021sgcn}. NL: evaluations only with non-linear trajectories, Diff: Performance difference (ADE/FDE, Unit: meter). \textbf{Bold}: Best.}
\label{tab:ablation_nonlinear}
\end{table}

\begin{table}[t]
    \Large
    \centering
    \resizebox{\linewidth}{!}{
\begin{tabular}{c c cccccc}
\toprule
Space & Noise & ETH & HOTEL & UNIV & ZARA1 & ZARA2 & AVG \\ \midrule

\multirow{4}{*}{Euclidean} & -    & 0.57 / 1.00 & 0.31 / 0.53 & 0.37 / 0.67 & 0.29 / 0.51 & 0.23 / 0.42 & 0.35 / 0.63 \\
                           & 0.02 & 0.58 / 1.01 & 0.35 / 0.59 & 0.39 / 0.69 & 0.30 / 0.54 & 0.27 / 0.48 & 0.38 / 0.66 \\
                           & 0.05 & 0.63 / 1.09 & 0.42 / 0.69 & 0.45 / 0.79 & 0.37 / 0.65 & 0.35 / 0.61 & 0.44 / 0.77 \\
                           & 0.10 & 0.75 / 1.28 & 0.60 / 1.00 & 0.61 / 1.05 & 0.56 / 0.98 & 0.53 / 0.89 & 0.61 / 1.04 \\ \cmidrule(lr){1-8}
\multicolumn{2}{c}{Diff.}  & \!-0.18\,/-0.29 & \!-0.29\,/-0.46 & \!-0.24\,/-0.38 & \!-0.27\,/-0.47 & \!-0.31\,/-0.47 & \!-0.26\,/-0.42 \\ \midrule
\multirow{4}{*}{\tbf{~~~\ETbold}} & -    & 0.36 / 0.57 & 0.13 / 0.21 & 0.24 / 0.43 & 0.20 / 0.35 & 0.15 / 0.26 & 0.22 / 0.36 \\
                                  & 0.02 & 0.37 / 0.58 & 0.14 / 0.22 & 0.25 / 0.44 & 0.21 / 0.37 & 0.17 / 0.28 & 0.23 / 0.38 \\
                                  & 0.05 & 0.41 / 0.62 & 0.17 / 0.25 & 0.29 / 0.48 & 0.26 / 0.42 & 0.20 / 0.32 & 0.26 / 0.41 \\
                                  & 0.10 & 0.47 / 0.68 & 0.25 / 0.33 & 0.36 / 0.55 & 0.36 / 0.53 & 0.27 / 0.39 & 0.34 / 0.50 \\ \cmidrule(lr){1-8}
\multicolumn{2}{c}{Diff.} & \!\tbf{-0.11}\,/\tbf{-0.11} & \!\tbf{-0.11}\,/\tbf{-0.12} & \!\tbf{-0.11}\,/\tbf{-0.12} & \!\tbf{-0.16}\,/\tbf{-0.18} & \!\tbf{-0.12}\,/\tbf{-0.13} & \!\tbf{-0.12}\,/\tbf{-0.13} \\
\bottomrule
\end{tabular}
}
\vspace{-1mm}
\caption{Ablation study on trajectory perturbation result of SGCN~\cite{Shi2021sgcn}. Noise: Standard deviation of Gaussian noise, Diff: Performance difference (ADE/FDE, Unit: meter). \textbf{Bold}:Best.}
\label{tab:ablation_noise}
\end{table}

\section{Conclusion}
In this work, we introduce a low-rank approximation-based trajectory descriptor trained in a data-driven manner to make a low-dimensional representation of pedestrian paths. 
While the existing architectures working in the Euclidean space suffer from the curse of dimensionality, we define a new operating space, the \ET~space, that unfolds highly-conjugated feature relations. We then cluster the coefficients of the \ET~descriptor coefficients on the training set, and utilize them as trajectory anchors. The architectures learn to refine this data-driven anchor to infer structurally-diverse trajectories that can cover all travelable paths. A variety of experiments demonstrate that it provides great applicability and stability, which can be applied to off-the-shelf trajectory forecasting models with consistent performance improvements on most public datasets.

\vspace{3mm}
\fontsize{8.4}{10}\selectfont{\noindent\textbf{Acknowledgement} This work is in part supported by the Institute of Information  $\&$ communications Technology Planning $\&$ Evaluation (IITP)  (No.2021-0-02068, Artificial Intelligence Innovation Hub), the National Research Foundation of Korea (NRF) (No.2020R1C1C1012635) grant funded by the Korea government (MSIT), GIST-MIT Research Collaboration grant funded by the GIST in 2023, and the Ministry of Trade, Industry and Energy (MOTIE) and Korea Institute of Advancement of Technology (KIAT) through the International Cooperative R\&D program: P0019782, Embedded AI Based fully autonomous driving software and Maas technology development.}

{\small
\bibliographystyle{ieee_fullname}
\bibliography{egbib}
}

\end{document}